%% file: iclr2023_workshop.tex
\documentclass{article} 
\usepackage{iclr2023_workshop,times}

\input{math_commands.tex}

\usepackage{hyperref}
\usepackage{url}
\usepackage{graphicx}
\usepackage{natbib}
\usepackage{subfig}
\usepackage{multirow}

\usepackage[normalem]{ulem}

\title{E($3$) Equivariant Graph Neural Networks for Particle-Based Fluid Mechanics}

\author{Artur Toshev, Gianluca Galletti, Stefan Adami \& Nikolaus Adams \\
Technical University of Munich, Chair of Aerodynamics and Fluid Mechanics \\
\texttt{\{artur.toshev,g.galletti\}@tum.de} \\
\And
Johannes Brandstetter \\
Microsoft Research AI4Science \\
}

\iclrfinalcopy 
\begin{document}

\maketitle

\newcommand\gnsacc{\tilde{\dot{v}}}
\newcommand\gnsout{\tilde{x}}
\newcommand\gnsstate{\tilde{\mathbf{X}}}
\newcommand\segnnacc{\tilde{\tilde{\dot{\mathbf{v}}}}}
\newcommand\segnnvel{\mathbf{v}}
\newcommand\segnnattr{\hat{a}}

\begin{abstract}
We contribute to the vastly growing field of machine learning for engineering systems by demonstrating that equivariant graph neural networks have the potential to learn more accurate dynamic-interaction models than their non-equivariant counterparts. We benchmark two well-studied fluid flow systems, namely the 3D decaying Taylor-Green vortex and the 3D reverse Poiseuille flow, and compare equivariant graph neural networks to their non-equivariant counterparts on different performance measures, such as kinetic energy or Sinkhorn distance. Such measures are typically used in engineering to validate numerical solvers. Our main findings are that while being rather slow to train and evaluate, equivariant models learn more physically accurate interactions. This indicates opportunities for future work towards coarse-grained models for turbulent flows, and generalization across system dynamics and parameters.
\end{abstract}

\begin{figure}[h]%
    \centering
    \subfloat[\centering]{{\includegraphics[height=4cm]{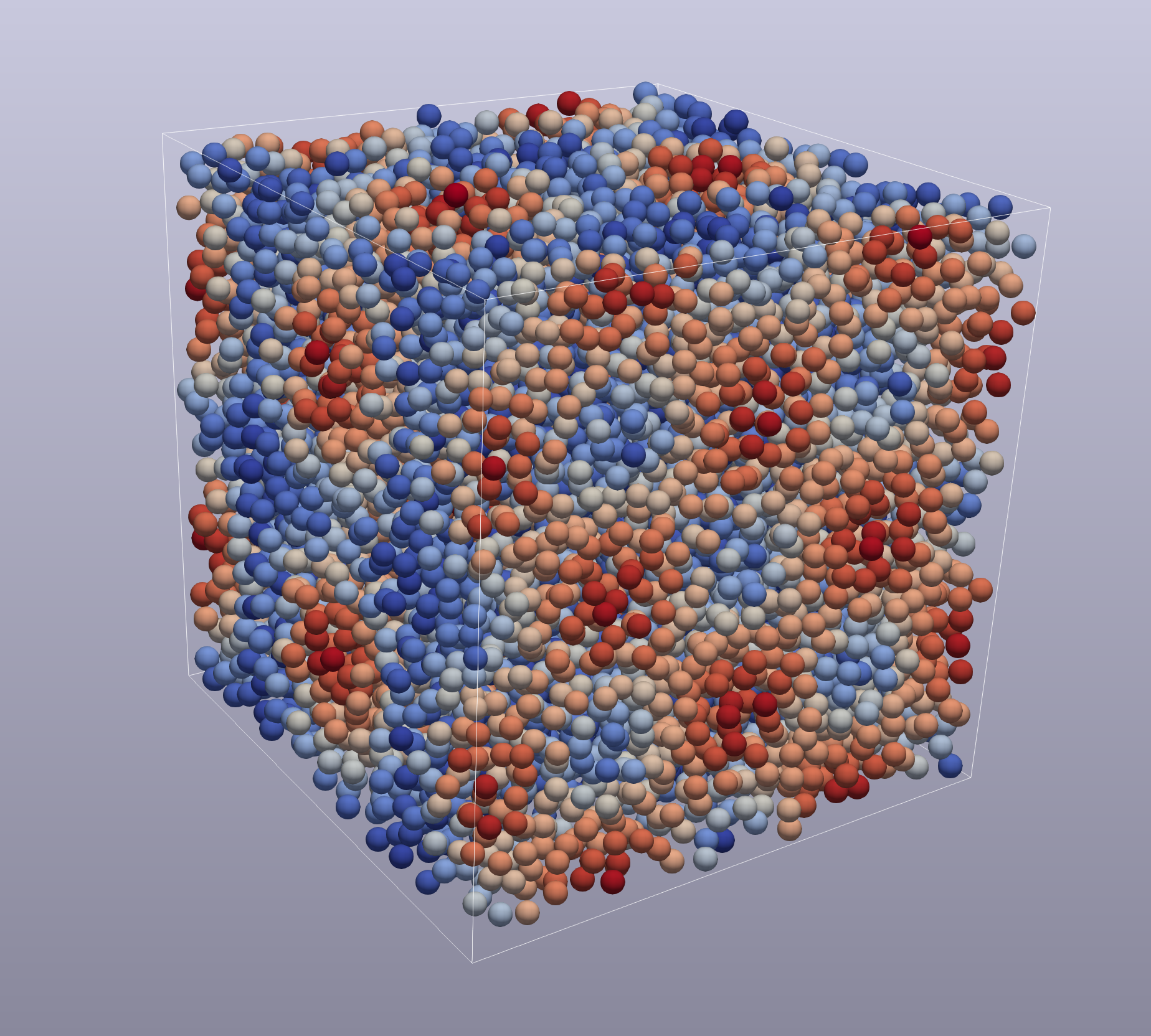} }}%
    \subfloat[\centering]{{\includegraphics[height=4cm]{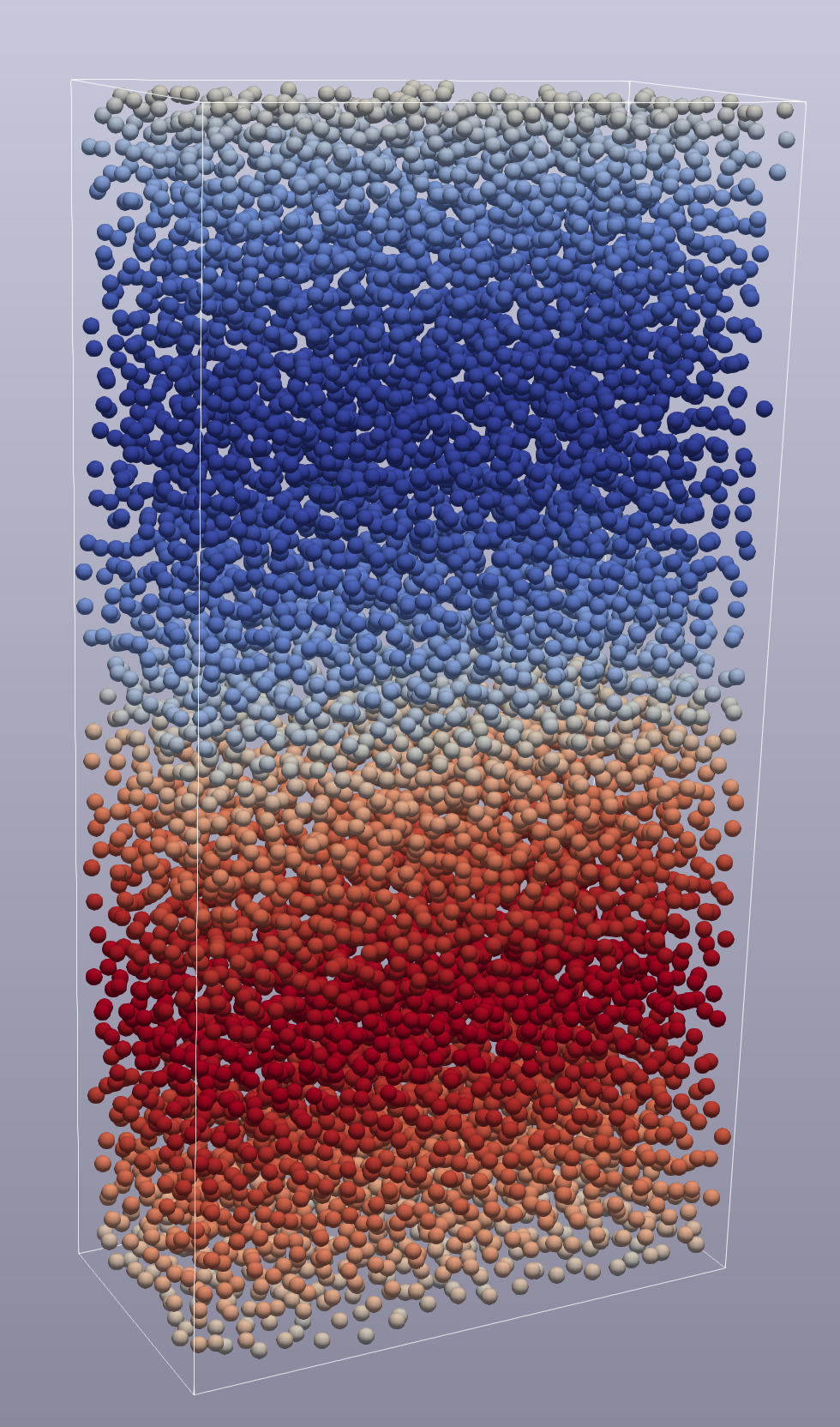} }}%
    \caption{Velocity magnitude of Taylor-Green vortex (a) and x-velocity of reverse Poiseuille (b).}%
    \label{fig:main}%
\end{figure}

\section{Particle-based fluid mechanics}

Navier-Stokes equations (NSE) are omnipresent in fluid mechanics, hydrodynamics or weather modeling.
However, for the majority of
problems, solutions are analytically intractable, and obtaining accurate predictions necessitates falling back to  numerical solution schemes.
Those can be split into two categories: grid/mesh-based (Eulerian description) and particle-based (Lagrangian description). 

\textbf{Smoothed Particle Hydrodynamics. } 
In this work, we investigate Lagrangian methods and more precisely the Smoothed Particle Hydrodynamics (SPH) approach, which was independently developed by \citet{gingold1977smoothed} and \citet{lucy1977numerical} to simulate astrophysical systems. Since then, SPH has established as the preferred approach in various applications ranging from free surfaces such as ocean waves \citep{violeau2016smoothed}, through fluid-structure interaction systems \citep{zhang2021multi}, to selective laser melting in additive manufacturing \citep{weirather2019smoothed}. The main idea behind SPH is to represent the fluid properties at discrete points in space and to use truncated radial interpolation kernel functions to approximate them at any arbitrary location. The kernel functions are used to estimate state statistics which define continuum-scale interactions between particles. The justification for truncating kernel support is the assumption of local interactions between particles. The resulting discretized equations are then integrated in time using numerical integration techniques like symplectic Euler by which the particle positions are updated.

To generate training data, we implemented our own SPH solver based on the transport velocity formulation by \citet{adami2013transport}, which promises a homogeneous particle distribution over the domain. We then selected two flow cases, both of which are well-known in the fluid mechanics community: the 3D laminar Taylor-Green Vortex and the 3D reverse Poiseuille Flow. 
We are planning to open-source the datasets in the near future.

\textbf{Taylor-Green Vortex. }
The Taylor-Green vortex system (TGV, see Figure~\ref{fig:main} (a)) with Reynolds number of $\text{Re}=100$ is neither laminar nor turbulent, i.e. there is no layering of the flow (typical for laminar flows), but also the small scales caused by vortex stretching do not lead to a fully developed energy cascade (typical for turbulent flows) \cite{brachet1984taylor}. The TGV has been extensively studied starting with \citet{taylor1937mechanism} and continuing all the way to \citet{sharma2019vorticity}. The TGV system is typically initialized with a velocity field given by
\begin{align}
    u = - \cos(k x) \cos(k y) \cos(k z) \ , \qquad
    v = \sin(k x) \cos(k y) \cos(k z) \ , \qquad
    w = 0 \ ,    
\end{align}
where $k$ is an integer multiple of $2 \pi$.
The TGV datasets used in this work consist of 8/2/2 trajectories for training/validation/testing,
where each trajectory comprises 8000 particles. Each trajectory spans 1s physical time and was simulated with $dt=0.001$ resulting in 1000 time steps per trajectory. The ultimate goal would be to learn the dynamics over much larger time steps than those taken by the numerical solver, but with this dataset we just want to demonstrate the applicability of learned approaches to reproducing numerical solver results. 

\textbf{Reverse Poiseuille Flow. }
The Poiseuille flow, i.e. laminar channel flow, is another well-studied flow case in fluid mechanics. However, channel flow requires the treatment of wall-boundary conditions, which is beyond the focus of this work. In this work, we therefore consider data obtained by reverse Poiseuille flow (RPF, see Figure~\ref{fig:main} (b)) \citep{fedosov2008reverse}, which essentially consists of two opposing streams in a fully periodic domain.
Those flows are exposed to opposite force fields, i.e., the upper and lower half are accelerated in negative $x$ direction and positive $x$ direction, respectively. 
Due to the fact that the flow is statistically stationary (the vertical velocity profile has a time-independent mean value), the RPF dataset consists of one long trajectory spanning 120s. The flow field is discretized by 8000 particles and simulated with $dt=0.001$, followed by subsampling at every 10th step.
Learning to directly predict every 10th state is what we call temporal coarse-graining. The resulting number of training/validation/testing instances is the same as for TGV, namely 8000/2000/2000.

\section{(Equivariant) graph network-based simulators}

We first formalize the task of autoregressively predicting the next state of a Lagrangian fluid mechanics simulation based on the notation from \citet{sanchez2020learning}. Let $X^t$ denote the state of a particle system at time $t$. One full trajectory of $K+1$ steps can be written as $\textbf{X}^{t_{0:K}}=(X^{t_0}, \ldots, X^{t_K})$. Each state $\textbf{X}^t$ is made up of $N$ particles, namely $\textbf{X}^t = (\textbf{x}_1^t, \textbf{x}_2^t, \dots \textbf{x}_N^t)$, where each $\textbf{x}_i$ is the state vector of the $i$-th particle. 
However, the inputs to the learned simulator can span multiple time instances. Each node $\textbf{x}_i^t$ can contain node-level information like the current position $\mathbf{p}^t_i$ and a time sequence of $H$ previous velocity vectors $\dot{\textbf{p}}^{t_{1+k-H:k}}$, as well as global features like the external force field $\textbf{f}_i$ in the reverse Poiseuille flow. To build the connectivity graph, we use an interaction radius of $\sim 1.5$ times the average interparticle distance. This results in around 10-20 one-hop neighbors.

\textbf{Graph Network-based Simulator.}
The Graph Network-based Simulator (GNS) framework~\citep{sanchez2020learning} is one of the most popular learned surrogates for engineering particle-based simulations. The main idea of the GNS model is to use the established encoder-processor-decoder architecture~\citep{battaglia2018relational} with a processor that stacks several message passing layers \citep{gilmer2017neural}. One major strength of the GNS model lies in its simplicity given that all its building blocks are simple MLPs. 
However, the performance of GNS 
when predicting long trajectories strongly depends on choosing the right amount of Gaussian noise to perturb input data.
Additionally, GNS and other non-equivariant models are less data-efficient~\citep{batzner20223}. For these reasons, we implement and tune GNS as a comparison baseline,
and use it as an inspiration for which setup, features, and hyperparameters to use for equivariant models.

\textbf{Steerable E(3)-equivariant Graph Neural Network. }
Steerable E(3)-equivariant Graph Neural Networks (SEGNNs)~\citep{brandstetter2021segnn} are an instance of E($3$)-equivariant GNNs, i.e., GNNs that are equivariant with respect to isometries of the Euclidean space (rotations, translations, and reflections). Most E($3$)-equivariant GNNs that are tailored towards molecular property prediction tasks, ~\citep{batzner20223,batatia2022mace} restrict the parametrization of the Clebsch-Gordan tensor products to an MLP-parameterized embedding of pairwise distances. In contrast, SEGNNs use general steerable node and edge attributes which can incorporate any kind of physical quantity,
and directly learn the weights of the Clebsch-Gordan tensor product. 
Indeed, extensions of methods such as NequIP \citep{batzner20223} towards general physical features would results in something akin to SEGNN.

Steerable attributes strongly impact the Clebsch-Gordan tensor products, and thus finding physically meaningful edge and node attributes is crucial for good performance. In particular, we chose edge attributes $\hat{a}_{ij} = V(\textbf{p}_{ij})$, where $V(\cdot)$ is the spherical harmonic embedding and $\textbf{p}_{ij}=\textbf{p}_{i}-\textbf{p}_{j}$ are the pairwise distances. We further choose node attributes $\hat{a}_{i} = V(\bar{\dot{\textbf{p}}}_{i}) + \sum_{k\in \mathcal{N}(i)} \hat{a}_{ik}$, where $\bar{\dot{\textbf{p}}}_{i}$ are averaged historical velocities and $\mathcal{N}(i)$ is the $i$-neighborhood.
As for node and edge features, we found that concatenated historical velocities for the nodes and pairwise displacements for the edges capture best the Navier-Stokes dynamics.



For training SEGNNs, we verified that adding Gaussian noise to the inputs~\citep{sanchez2020learning} indeed significantly improves performance. We further found that explicitly concatenating the external force vector $\mathbf{f}_i$ to the node features boosts performance in the RPF case. However, adding $\mathbf{f}_i$ to the node attributes $\hat{a}_{i}\prime = V(\mathbf{f}_i) + V(\bar{\dot{\textbf{p}}}_{i}) + \sum_{k\in \mathcal{N}(i)} \hat{a}_{ik}$ does not improve performance.

Other models, like EGNN by \citet{satorras2021en}, achieve equivariance by working with invariant messages, but it does not allow the same flexibility in terms of features. On a slightly more distant note, there has been a rapid raise in physics-informed machine learning~\citep{raissi2019physics} and operator learning~\citep{li2021fourier}, where functions or surrogates are learned in an Eulerian (grid-based) way. SEGNN is a sound choice for Lagrangian fluid mechanics problems since it is designed to work directly with vectorial information and particles. 


\section{Results}

The task we train on is the autoregressive prediction of accelerations $\ddot{\mathbf{p}}$ given the current position $\mathbf{p}_i$ and $H=5$ past velocities of the particles. We measured the performance of the GNS and the SEGNN models in four aspects when evaluating on the test dataset: (i) \textit{Mean-squared error} (MSE) of particle positions $\text{MSE}_p$ when rolling out a trajectory over 100 time steps (1 physical second for both flow cases). This is also the validation loss during training. (ii) \textit{Sinkhorn distance}, as an optimal transport distance measure between particle distributions. Lower values indicate that the particle distribution is closer to the reference one. (iii) \textit{Kinetic energy} $E_{kin}$ ($=0.5 m v^2$) as a global measure of physical behavior.
\begin{table}[]
\centering
\setlength{\tabcolsep}{8pt}
\renewcommand{\arraystretch}{1.2}

\caption{Performance measures on the Taylor-Green vortex and reverse Poiseuille flow.
The Sinkhorn distance is averaged over test rollouts, the inference time is obtained for one rollout step of 8000 particles. 
}
\begin{tabular}{cccc|ccc}
 & \multicolumn{3}{c|}{Taylor-Green vortex} & \multicolumn{3}{c}{Reverse Poiseuille flow} \\ \cline{2-7}
\multicolumn{1}{c}{} & \multicolumn{1}{c}{SEGNN} & \multicolumn{1}{c}{GNS} & SPH & \multicolumn{1}{c}{SEGNN} & \multicolumn{1}{c}{GNS} & SPH \\ \hline
\multicolumn{1}{c|}{$\text{MSE}_{\textbf{p}}$} & \multicolumn{1}{c}{7.7e-5} & \multicolumn{1}{c}{1.3e-4} & - & \multicolumn{1}{c}{7.7e-3} & \multicolumn{1}{c}{8.0e-3} & - \\
\multicolumn{1}{c|}{$\text{MSE}_{E_{kin}}$} & \multicolumn{1}{c}{5.3e-5} & \multicolumn{1}{c}{1.3e-4} & - & \multicolumn{1}{c}{2.8e-1} & \multicolumn{1}{c}{3.0e-1} & - \\
\multicolumn{1}{c|}{$\overline{\text{Sinkhorn}}$} & \multicolumn{1}{c}{1.3e-7} & \multicolumn{1}{c}{1.1e-7} & - & \multicolumn{1}{c}{7.8e-8} & \multicolumn{1}{c}{1.9e-6} & - \\
\multicolumn{1}{c|}{Time [ms]} & \multicolumn{1}{c}{290} & \multicolumn{1}{c}{32} & 9.7 & \multicolumn{1}{c}{180} & \multicolumn{1}{c}{33} & 110 \\
\multicolumn{1}{c|}{$\#$ params} & \multicolumn{1}{c}{720k} & \multicolumn{1}{c}{630k} & - & \multicolumn{1}{c}{180k} & \multicolumn{1}{c}{630k} & - 

\end{tabular}
\label{table:comparison}
\end{table}
Performance comparisons are summarized in Table \ref{table:comparison}. GNS and SEGNN models have roughly the same number of parameters for Taylor-Green (both have 5 layers and 128-dim features), whereas for reverse Poiseuille SEGNN has three times less parameters than GNS (SEGNN has 64-dim features). Looking at the timing in Table \ref{table:comparison}, equivariant models of similar size are one order of magnitude slower than non-equivariant ones. This is a known result and is related to the constraint of how the Clebsch-Gordan tensor product can be implemented on accelerators like GPUs.

\begin{figure}[t]%
    \centering
    \subfloat[\centering]{{\includegraphics[height=6.5cm]{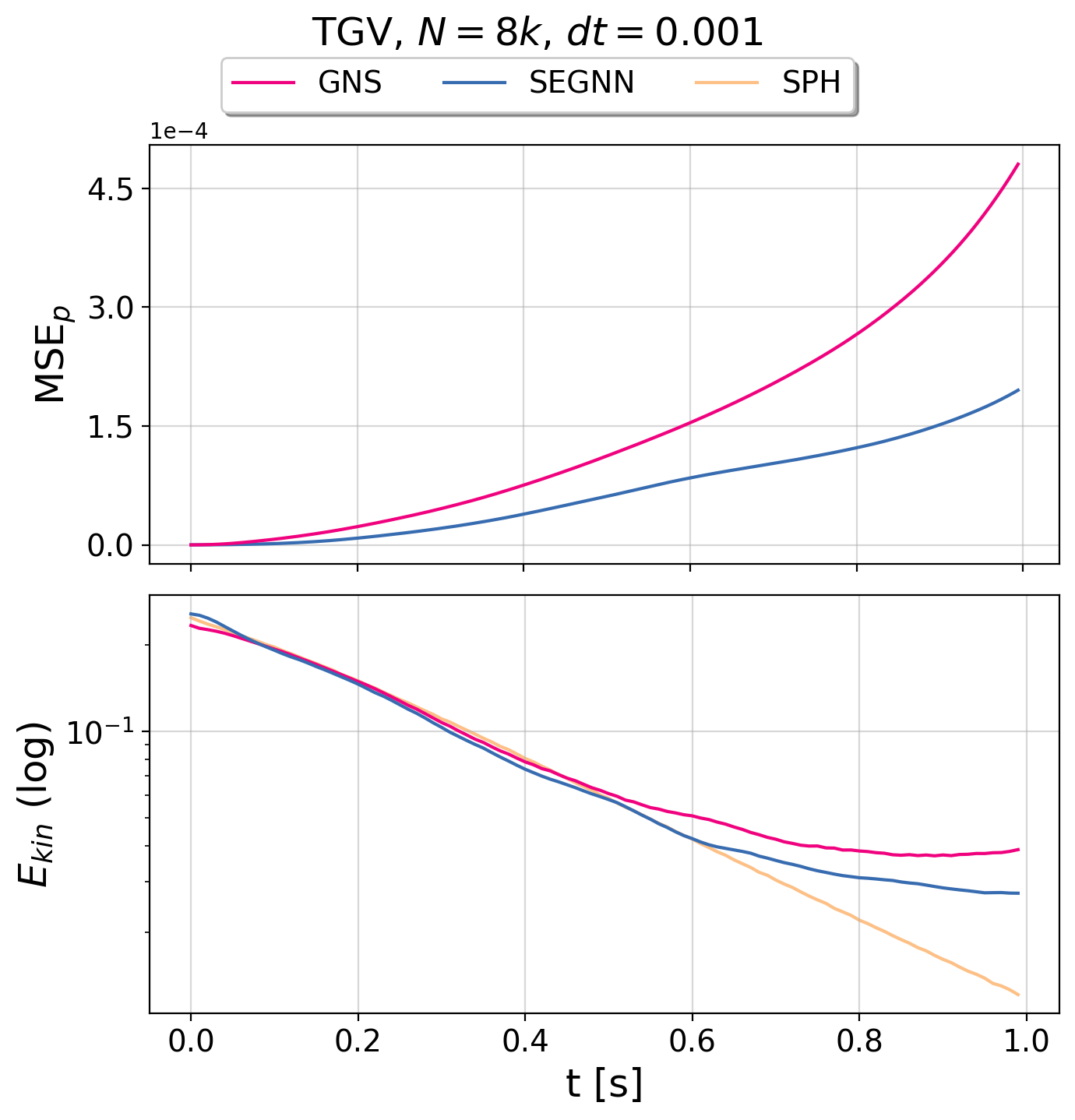} }}%
    \subfloat[\centering]{{\includegraphics[height=6.5cm]{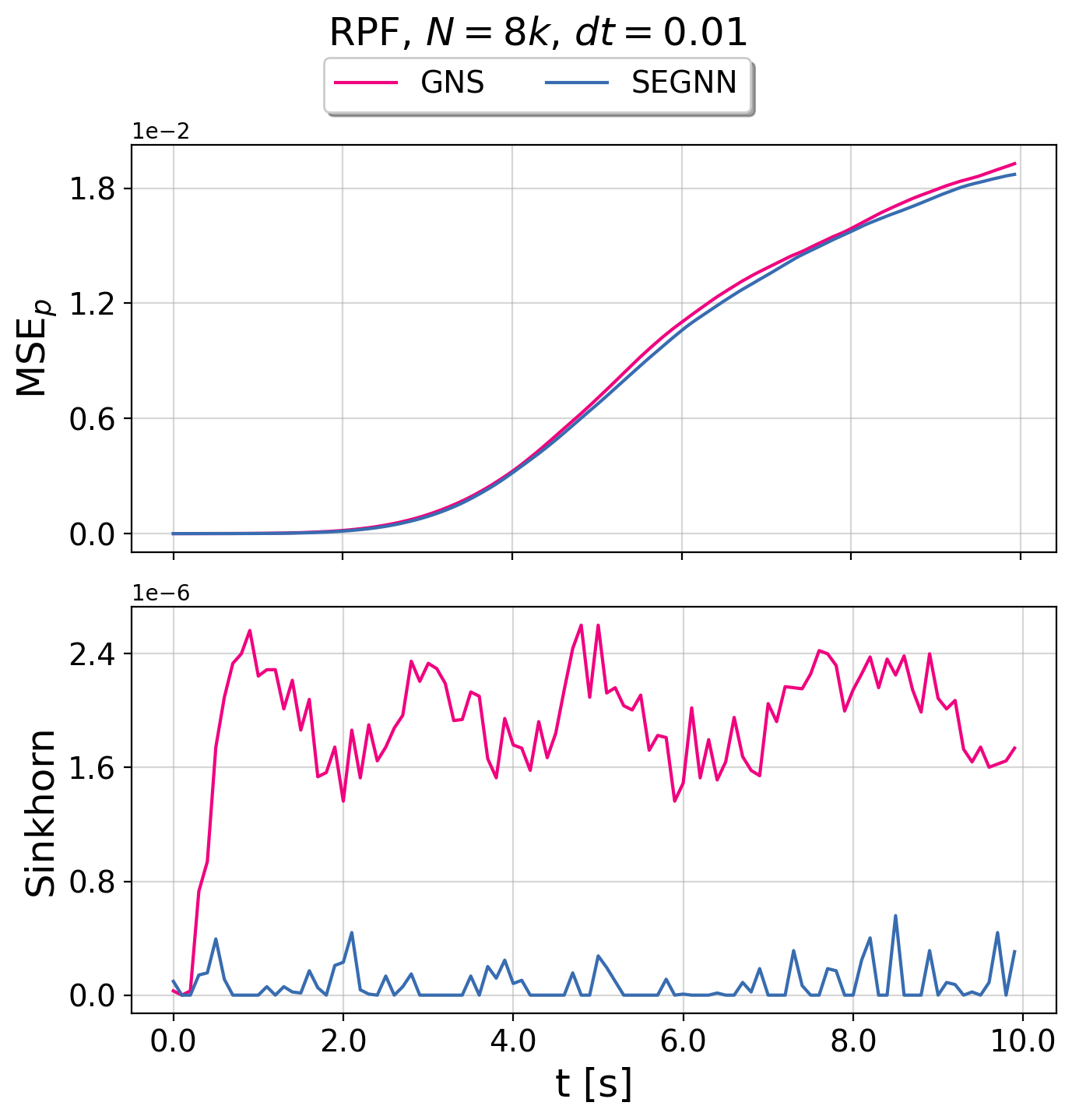} }}%
    \caption{Taylor-Green vortex (a) and reverse Poiseuille (b) performance evolution.}%
    \label{fig:metrics}%
\end{figure}

\textbf{Taylor-Green Vortex. } 
One of the major challenges of the Taylor-Green dataset are the varying input and output scales throughout a trajectory, by up to one order of magnitude. Consequently, this results in the larger importance of first-time steps in the loss even after data normalization. 
Figure \ref{fig:metrics} (a) summarizes the most important performance properties of the Taylor-Green vortex experiment.
In general, both models match the ground truth kinetic energy well, but GNS drifts away from the reference SPH curve earlier. Both learned solvers, seem to preserve larger system velocities resulting in higher $E_{kin}$. The rollout MSE for this case matches the behavior seen in $E_{kin}$.

\textbf{Reverse Poiseuille Flow. }
The challenge of the reverse Poiseuille case lies in the different velocity scales between the main flow direction ($x$-axis) and the $y$ and $z$ components of the velocity. Although such unbalanced velocities are used as inputs, target accelerations in $x$-, $y$-, and $z$-direction all underlie similar distributions. This, combined with temporal coarsening makes the problem sensitive to input deviations.
Figure \ref{fig:metrics} (b) shows that SEGNN reproduces the particle distribution almost perfectly, whereas GNS shows signs of particle clustering, resulting in a larger Sinkhorn distance. Interestingly, the shear layers in-between the inverted flows (around planes $y=\{0, 1, 2\}$) seem to have the largest deviation from the ground truth, which could be source of clusters, see Figure \ref{fig:rpf_error}.

\section{Future Work}
In this work, we demonstrate that equivariant models are well suited to capture underlying physics properties of particle-based fluid mechanics systems. 
Natural future steps are enforcing physical behaviors such as homogeneous particle distributions, and including recent developments for neural PDE training into the training procedure of~\citet{sanchez2020learning}. The latter include e.g., the push-forward trick and temporal bundling~\citep{brandstetter2022message}.
One major weakness of recursively applied solvers, which these strategies aim to mitigate, is error accumulation, which in most cases leads to out-of-distribution states, and consequently unphysical behavior after several rollout steps. 
We conjecture that together with such extensions equivariant models offer a promising direction to tackle some of the long-standing problems in fluid mechanics, such as the learning of coarse-grained representations of turbulent flow problems, e.g. Taylor-Green \citep{brachet1984taylor}, or learning the multi-resolution dynamics of NSE problems \citep{hu2017consistent}.


\bibliography{iclr2023_workshop}
\bibliographystyle{iclr2023_workshop}

\appendix
\section{Appendix}

\textbf{Reverse Poiseuille plots. }

\begin{figure}[h]%
    \centering
    \includegraphics[height=12cm,trim={0 0 0 1cm},clip]{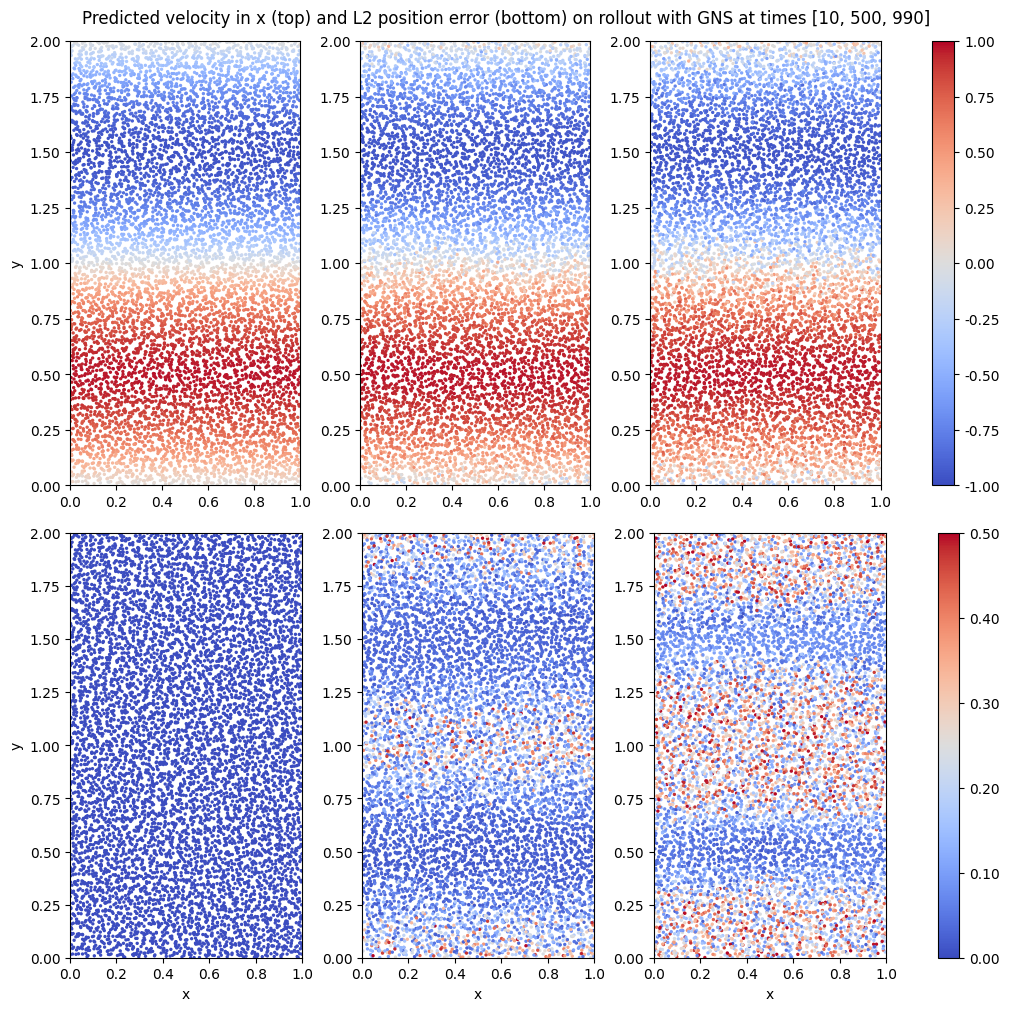} %
    \caption{Reverse Poiseuille x-y view of velocity fields (top) and position error (bottom) at time steps [10, 500, 990] (left to right) of the test rollout.}%
    \label{fig:rpf_error}%
\end{figure}

\end{document}

%% file: math_commands.tex
\usepackage{amsmath,amsfonts,bm}

\def\eqref#1{equation~\ref{#1}}

\def\1{\bm{1}}

\DeclareMathAlphabet{\mathsfit}{\encodingdefault}{\sfdefault}{m}{sl}
\SetMathAlphabet{\mathsfit}{bold}{\encodingdefault}{\sfdefault}{bx}{n}

